\documentclass[sn-mathphys-num]{sn-jnl}% Math and Physical Sciences Numbered Reference Style 
\usepackage{graphicx}%
\usepackage{multirow}%
\usepackage{amssymb,amsfonts}%
\usepackage{amsthm}%
\usepackage[fleqn]{amsmath}
\usepackage{mathrsfs}%
\usepackage[title]{appendix}%
\usepackage{xcolor}%
\usepackage{textcomp}%
\usepackage{manyfoot}%
\usepackage{booktabs}%
\usepackage{algorithm}%
\usepackage{algorithmicx}%
\usepackage{algpseudocode}%
\usepackage{listings}%
\usepackage{rotating}  % Add in your preamble
\usepackage[normalem]{ulem}
\usepackage{amsmath}
\theoremstyle{thmstyleone}%
\theoremstyle{thmstyletwo}%

\theoremstyle{thmstylethree}%

\begin{document}

% \title[Article Title]{A Light Gradient Residual Encoder-Decoder Network for Multimodal Image Fusion}

\title[Article Title]{SWIR-LightFusion: Multi-spectral Semantic Fusion of Synthetic SWIR with {Thermal} IR {(LWIR/MWIR)} and RGB}


\author[1,3]{\fnm{Muhammad Ishfaq} \sur{Hussain}}\email{\{ishfaqhussain}

\author[1]{\fnm{Ma Van} \sur{Linh}}\email{linh.mavan}

\author[1]{\fnm{Zubia} \sur{Naz}}\email{zubianaz}
\author[1]{\fnm{Unse} \sur{Fatima}}\email{unse.fatima}
\author[2]{\fnm{Yeongmin} \sur{Ko}}\email{koyeongmin}
\author[1]{\fnm{Moongu} \sur{Jeon}}\email{mgjeon\}@gist.ac.kr}


\affil[1]{%\orgdiv{School of Electrical Engineering and Computer Science}, %\orgname{Gwangju Institute of Science and Technology}, 
\orgname{GIST}, 
%\orgaddress{\city{Gwangju}, \postcode{61005}, \state{Buk-gu}, \country{Rep.of Korea}}}
\orgaddress{\city{Gwangju}, \country{Rep.of Korea}}}

\affil[2]{\orgdiv{} \orgname{Kyungpook National University}, \orgaddress{\city{Daegu}, \country{Rep.of Korea}}}%\postcode{41566}, 

\affil[3]{%\orgdiv{Korea Institute of Science and Technology Information}, 
\orgdiv{KISTI}, \orgaddress{\city{Daejeon},  \country{Rep.of Korea}}}%\postcode{34141},

\abstract{
Enhancing scene understanding in adverse visibility conditions remains a critical challenge for surveillance and autonomous navigation systems. Conventional imaging modalities, such as RGB and {thermal} infrared (MWIR / LWIR), when fused, often struggle to deliver comprehensive scene information, particularly under conditions of atmospheric interference or inadequate illumination. To address these limitations, Short-Wave Infrared (SWIR) imaging has emerged as a promising modality due to its ability to penetrate atmospheric disturbances and differentiate materials with improved clarity. However, the advancement and widespread implementation of SWIR-based systems face significant hurdles, primarily due to the scarcity of publicly accessible SWIR datasets. In response to this challenge, our research introduces an approach to synthetically generate SWIR{-like structural/contrast cues (without claiming spectral reproduction)} images from existing {L}WIR data using advanced contrast enhancement techniques. We {then} propose a multimodal fusion framework integrating synthetic SWIR, {L}WIR, and RGB modalities, employing an optimized encoder-decoder neural network architecture {with modality-specific encoders and a softmax-gated fusion head}. Comprehensive experiments on public {RGB-LWIR benchmarks (M3FD, TNO, CAMEL, MSRS, RoadScene) and an additional private real RGB-MWIR-SWIR dataset} demonstrate that our synthetic-SWIR-enhanced fusion framework improves fused-image quality (contrast, edge definition, structural fidelity) while maintaining real-time performance. {We also add fair trimodal baselines (LP, LatLRR, GFF) and cascaded trimodal variants of U2Fusion/SwinFusion under a unified protocol.}The outcomes highlight substantial potential for real-world applications in surveillance and autonomous systems. Details of synthetic SWIR generation and fusion methodology will be publicly available at \href{https://github.com/MI-Hussain/SynthSWIRNet\_2}{https://github.com/MI-Hussain/LightFusion\_2}.}


\maketitle

\section{Introduction}\label{sec1}
With the rapid advancement and widespread deployment of diverse sensing technologies, multimodal image fusion has become a crucial capability across numerous application domains. Visible spectrum (RGB) and infrared (IR) imaging, particularly Mid-Wave \&{Long-wave} Infrared (MWIR{/LWIR}), have received considerable attention due to their complementary imaging properties, each offering unique advantages. This fusion can enhance object detection and recognition under varying environmental conditions such as low light or fog, where RGB provides detailed color information and (MWIR{/LWIR}) detects thermal signatures \cite{liu2022target, tang2022piafusion, zhang2021image}. This integrated approach broadens application possibilities, making it valuable in fields like surveillance, automotive safety, and healthcare. RGB imagery excels in capturing detailed texture and vibrant colors, but its effectiveness sharply declines under poor lighting conditions \cite{hussain2022drivable}. In contrast, IR imaging effectively highlights thermal characteristics of objects, offering enhanced visibility in darkness or challenging atmospheric conditions but typically lacking fine texture detail \cite{tang2022image, tang2023rethinking}.

Recent research has increasingly focused on enhancing image fusion techniques at the feature level to improve performance in downstream tasks, such as object detection and tracking \cite{tang2022piafusion}. Traditionally, RGB-IR object detection approaches utilize "late fusion" techniques, where modality-specific features are combined or concatenated, but such methods often fall short in adequately leveraging complementary information, thus limiting their performance \cite{liu2024multi}. Other strategies, like "halfway fusion," attempt to integrate interactive modules between modalities to strengthen the fusion quality, yet remain susceptible to modality-specific noise and fail to deliver fully integrated complementary information \cite{liu2024multi, hussain2022exploring, hussain2023artificial}. Inspired by cognitive theories such as Treisman's "Attenuation Theory," which advocates a hierarchical filtering mechanism to suppress irrelevant information, researchers introduced the Redundant Spectrum Removal (RSR) module to eliminate extraneous frequency information, followed by selective fine-grained feature fusion \cite{zhao2024removal}. To further enrich semantic understanding and adaptability in practical visual scenarios, semantic-aware fusion frameworks such as SeAFusion have emerged, integrating segmentation models to enhance semantic detail within fused outputs \cite{tang2022image}. However, these approaches may limit versatility across diverse models and struggle under extreme environmental conditions \cite{9216363}. In parallel, feature-level fusion techniques, processing multimodal features directly without producing explicit fused imagery, have gained prominence, particularly in multi-object tracking applications \cite{van2024visual, linh2024inffus}. Nonetheless, current feature-level fusion strategies typically require extensive redesigns for compatibility with modern deep learning architectures, such as Transformers or ConvNeXt, and have not fully realized the potential benefits available through comprehensive image-level fusion \cite{ranipa2024novel}. 

Addressing these gaps in our previous research \cite{hussain2024light}, we proposed a three-band sensor fusion framework utilizing grayscale images converted from RGB imagery due to the unavailability of SWIR images. Although this grayscale-based fusion provided enhanced visual clarity, its performance was limited due to the inherent constraints of grayscale representation compared to the advantages offered by real SWIR data. However, grayscale could not cover the in-depth analysis to cover the gap of SWIR \cite{kumar2024mwirstd}, this research introduces an innovative multimodal fusion network designed to combine RGB, {LW}IR, and a synthetically generated Short-Wave Infrared (SWIR) modality, specifically created to overcome the scarcity of open-source SWIR datasets. SWIR imaging provides critical benefits including superior penetration through atmospheric interference, greater contrast, and enhanced differentiation of materials, making it highly valuable for improved visual understanding. SWIR is less sensitive to temperature variations compared to {LW}IR, enabling it to provide stable imaging in fluctuating thermal environments, which is crucial for outdoor surveillance and agricultural monitoring. Additionally, the ability of SWIR to differentiate between various materials based on their spectral signatures is beneficial in various applications. Our synthetic SWIR generation leverages advanced contrast enhancement techniques applied to {LW}IR imagery, significantly enriching the available visual information. The proposed fusion architecture employs a lightweight gradient residual encoder-decoder framework optimized for real-time processing, independently extracting modality-specific semantic features and efficiently integrating them for enhanced scene perception and understanding.

The significant contributions of this research include:
\begin{itemize}
\item Introduced synthetic Short-Wave Infrared (SWIR{-like}) images as a third modality, significantly enhancing multimodal fusion systems by providing additional spectral information that improves detection capabilities in challenging conditions.\\
\item The development of an efficient, semantic-aware encoder-decoder network tailored for multimodal fusion involving RGB, {L}WIR, and synthetic SWIR images. {Additional experiments are performed with private real dataset contains SWIR, MWIR and RGB dataset to show the importance of proposed network}\\
\item Comprehensive evaluation of the proposed approach on multiple benchmarks including the M3FD \cite{liu2022target}, TNO \cite{toet2017tno}, MSRS \cite{tang2022piafusion}, Camel \cite{gebhardt2018camel} and RoadScene \cite{xu2020aaai} datasets, affirming its effectiveness and practical applicability in challenging environmental scenarios.
\end{itemize}

\section{Related Work}\label{sec2}

Traditional multimodal image fusion techniques have primarily focused on feature extraction and integration using various classical methodologies. Techniques such as multi-scale transforms—including Laplacian pyramids and discrete wavelet transforms—alongside sparse representation methods and subspace-based techniques such as Principal Component Analysis (PCA) have been widely explored \cite{hu2022robust}. Further advancements include optimization-driven and hybrid frameworks that combine several traditional approaches for improved performance. Despite their utility, these methods generally struggle to capture complex inter-modal relationships adequately.

The introduction of deep learning has significantly transformed the landscape of image fusion. Autoencoder architectures have gained traction by learning to reconstruct high-quality fusion outputs from modality-specific features, leveraging convolutional layers and dense blocks to merge complex information effectively. Similarly, convolutional neural networks (CNNs) provide powerful implicit feature extraction capabilities, effectively automating feature selection and aggregation, thereby surpassing manually designed fusion strategies \cite{schlemper2017deep}. Additionally, Generative Adversarial Networks (GANs) have demonstrated potential in image fusion tasks by employing adversarial training paradigms to enhance textural richness and edge sharpness in fused outputs, despite challenges such as instability and mode collapse \cite{fu2021image}.

The fusion of infrared (IR) and visible (RGB) imaging modalities, termed Infrared-Visible Image Fusion (IVIF), has been an active research area, addressing key tasks such as object detection and tracking \cite{liu2022target, tang2022piafusion}. Initial approaches, such as the work by Liu et al. \cite{liu2022target}, integrated pre-trained CNNs to learn modality-specific features and fusion weights. Recent studies, such as those leveraging residual fusion networks, have aimed to achieve more consistent structural representations by unifying features into a shared embedding space \cite{li2024residual}. FusionGAN \cite{ma2019fusiongan} further enhanced IVIF by adversarially training to merge texture-rich visible images with infrared data; however, it occasionally failed to retain essential IR information clearly. Moreover, extensive efforts have been invested in developing reliable benchmarks for IVIF research, such as the TNO Image Fusion, OSU Color-Thermal, RoadScene \cite{xu2020aaai}, and Multispectral datasets, each offering diverse scenarios and specific evaluation challenges \cite{liu2022target, toet2017tno, gebhardt2018camel}. Despite these significant contributions, current IVIF methodologies often inadequately represent the unique properties of distinct spectral bands, limiting their effectiveness across varied application scenarios.

Our previous research \cite{hussain2024light} introduced a novel three-band fusion approach employing grayscale images derived from RGB inputs to complement IR and RGB data, addressing real-time deployment constraints by enhancing visual contrast and detail. Nevertheless, grayscale representations intrinsically lack the spectral specificity and robustness of actual Short-Wave Infrared (SWIR) imagery. The lack of publicly available datasets featuring authentic SWIR and Infrared ({LWIR /}MWIR) imagery further restricts advancements in multimodal fusion technologies. To mitigate this, Kumar et al. \cite{kumar2024mwirstd} recently proposed the MWIR Small Target Detection Dataset (MWIRSTD), providing genuinely captured MWIR imagery with diverse small targets in realistic environmental contexts, significantly improving the robustness of models for detection and tracking applications. Recognizing these critical gaps, our current research proposes an innovative multimodal fusion framework utilizing synthetic SWIR images generated from LWIR imagery, specifically addressing the limited availability of authentic SWIR datasets. Our proposed fusion method employs an advanced semantic-aware encoder-decoder architecture capable of independently extracting and integrating features from RGB, ({LWIR /}MWIR), and synthetic SWIR modalities, significantly enhancing visual clarity and semantic robustness for practical real-time applications.

\section{Proposed Methodology}
In this paper, we introduce SWIR-LightFusion, a novel and computationally efficient encoder-decoder neural network designed specifically for robust multimodal image fusion. Building upon our previously introduced LightFusion framework \cite{hussain2024light}, which independently processed RGB, IR, and grayscale modalities, the current approach significantly extends this architecture by incorporating synthetically generated Short-Wave Infrared (SWIR) imagery from Infrared ({LWIR /}MWIR) data. This innovation is particularly valuable given the limited availability of authentic SWIR datasets and exploits SWIR's unique advantages in material differentiation and atmospheric penetration.

\subsection{Network Architecture}

\begin{figure*}[ht!]
    \centering
    \includegraphics[width=0.99\textwidth]{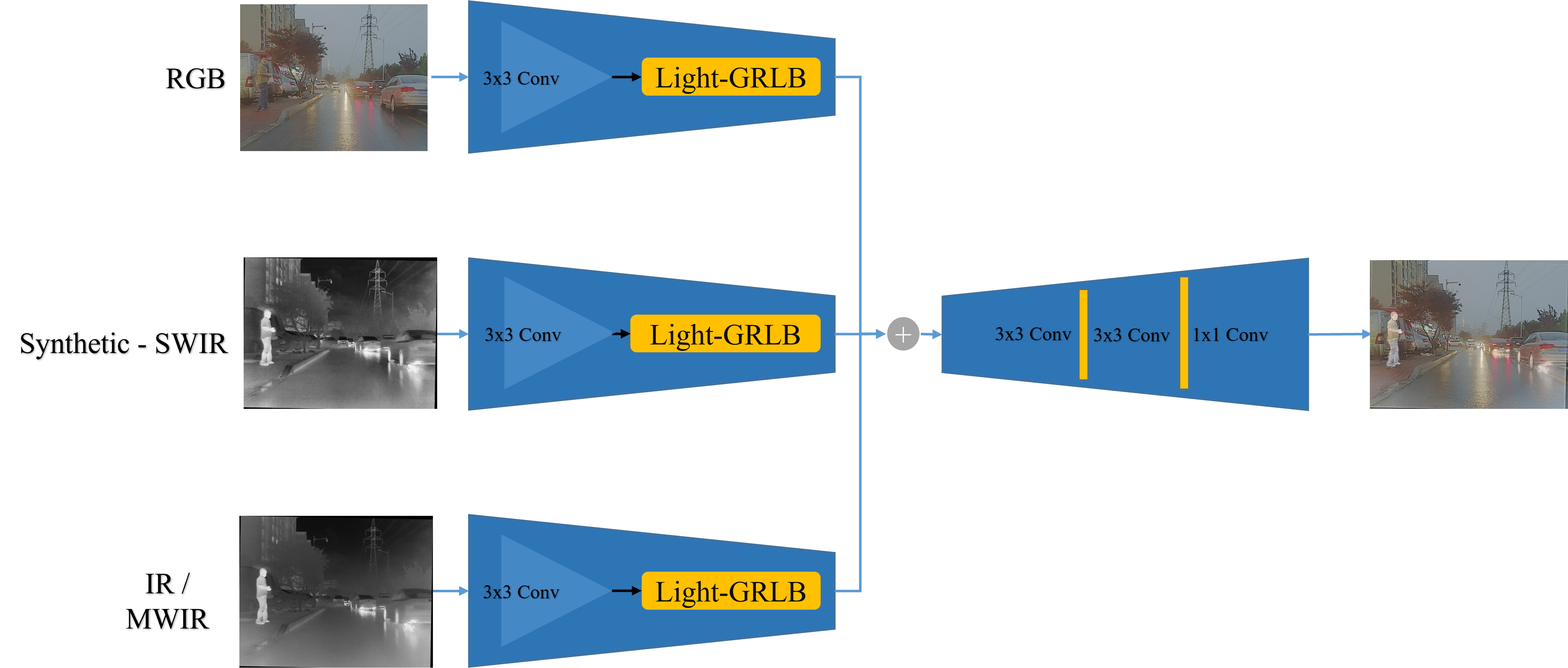}
    \caption{The overall structure of the proposed SWIR-LightFusion network. RGB, MWIR, and synthetically generated SWIR images are independently processed by dedicated encoder modules consisting of lightweight gradient residual blocks (Light-GRLB). Extracted features are then seamlessly integrated within the fusion layer, subsequently reconstructed by the decoder module to generate a high-quality fused output image.}
    \label{fig:mesh1}
\end{figure*}

The overall structure of SWIR-LightFusion, depicted in Fig. \ref{fig:mesh1}, comprises three primary modality-specific encoders that independently handle RGB, MWIR, and synthetic SWIR bands. Each encoder module utilizes a specialized lightweight gradient residual block (Light-GRLB), which employs gradient-based recurrent neural networks (RNNs) for efficient yet detailed feature extraction. These modules effectively capture subtle temporal and spatial dependencies intrinsic to each modality, thereby enhancing the richness and relevance of extracted features.

\noindent\textbf{Synthetic SWIR {SynSWIR} Generation}: SWIR images capture light in the short-wave infrared spectrum, typically defined as wavelengths between 0.9--1.7~$\mu$m, though sometimes extending to 0.7--2.5~$\mu$m\footnote{\url{https://www.edmundoptics.com/knowledge-center/application-notes/imaging/what-is-swir/}}. Due to the scarcity of open-source SWIR datasets, we introduce a synthetic SWIR generation process using Contrast Limited Adaptive Histogram Equalization (CLAHE) \cite{musa2018review} applied directly to {L}WIR images. This approach significantly enhances the contrast and visual clarity of {L}WIR images, closely emulating genuine SWIR data characteristics. This preprocessing step crucially expands the available data for fusion, augmenting the network's capability to interpret complex visual scenarios. 
{Let $I:\Omega\!\to\!\mathbb{R}$ denote the thermal image (LWIR/MWIR). CLAHE computes, for each pixel $x\in\Omega$, a locally monotone mapping $T_{\Omega(x)}$ from the clipped local histogram, yielding
\[
J(x)=T_{\Omega(x)}\!\big(I(x)\big),\quad T'_{\Omega(x)}(t)\ge 0.
\]
Hence edge polarity is preserved ($\mathrm{sign}(\nabla J)=\mathrm{sign}(\nabla I)$ a.e.) and edge magnitude is locally reweighted ($\|\nabla J\|=T'_{\Omega(x)}(I)\,\|\nabla I\|$), amplifying weak but semantically important boundaries while clipping suppresses over-enhancement in homogeneous regions. We use $J$ as a \emph{SWIR-like} proxy (SynSWIR) that concentrates high-frequency, material-boundary cues that SWIR typically makes salient. This proxy is \emph{not} intended to be spectrally faithful; its role is to provide an additional, complementary view for robust, real-time fusion when real SWIR is unavailable.}
\\
{On the rationale of SynSWIR Proposition: For any differentiable, monotone $T$, $J=T\!\circ\! I$ preserves edge polarity and scales edge magnitude by $T'(I)$ almost everywhere. Moreover, CLAHE's local monotonicity implies that, restricted to an image graph with $L^1$-bounded variation, the number of critical points does not increase, helping prevent spurious texture creation while enhancing contrast around existing transitions. \emph{Sketch.} By the chain rule $\nabla J=T'(I)\nabla I$; since $T'\!\ge 0$, gradient direction is unchanged while magnitude is locally reweighted. CLAHE's clip limit bounds $T'$ and thus noise amplification.
\\
Quantifying similarity to real SWIR: In Sec.~4.7 we report that trimodal fusion with SynSWIR reaches over 90\% of the gains of real SWIR across metrics on our real three-band set (Table~7). To further assess distributional similarity, we compute the Jensen--Shannon divergence between edge-histogram distributions of SynSWIR and real SWIR inside Canny edge masks; lower JS indicates closer structural statistics. (Results added to App.~B.)}

\begin{figure*}[ht]
    \centering
 \includegraphics[width=12cm, height=2.5cm]{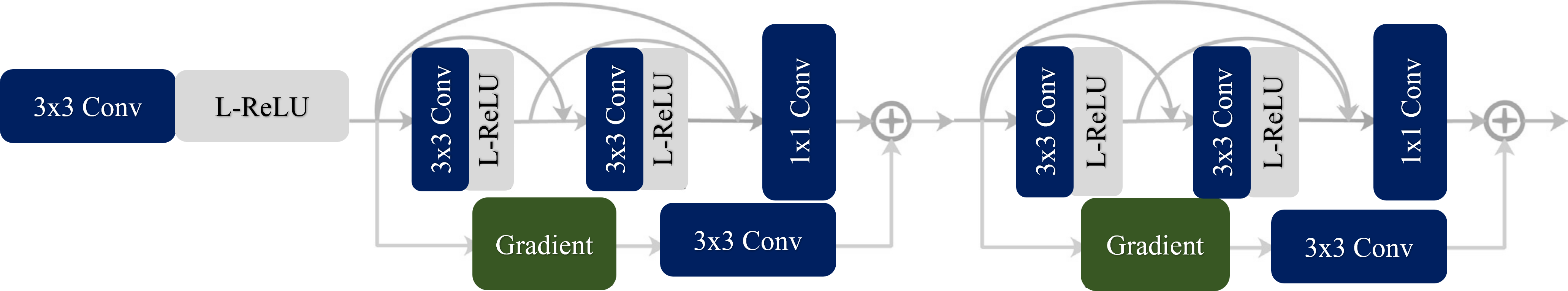}
    \caption{An in-depth explanation of Gradient Residual-based Convolutional Layers \cite{hussain2024light}, illustrating the integration of pointwise (1x1) convolutions to reduce computational complexity and make the network lightweight while preserving essential gradient features.}
    \label{fig:mesh2}
\end{figure*}

\noindent\textbf{Encoder Module}: Each modality-specific encoder is constructed from lightweight gradient residual blocks (Light-GRLB) as shown in Fig. \ref{fig:mesh2}, which consist of two interconnected convolutional blocks designed for low computational overhead while maximizing feature extraction efficiency. Each block consists of 3x3 convolutional layers integrated with LReLU activation. The block concludes with a 1x1 convolutional layer aimed at incrementally refining and synthesizing the fused image while being computationally efficient. Dense skip connections facilitate a direct and continuous flow of information between all layers across each block, significantly enhancing feature propagation and preserving gradient strength throughout the network. By enabling each layer to receive inputs from all preceding layers and to contribute to all subsequent layers, the proposed approach addresses vanishing gradients and ineffective feature utilization issues effectively. Gradient-based operation is used to extract detailed features through recurrent neural networks (RNNs). It effectively accentuates subtle differences in the image's textures and edges, bringing forth greater detail by assessing changes in intensity between neighboring pixels. These encoders independently analyze RGB, MWIR, and synthetic SWIR inputs, ensuring precise modality-specific feature extraction and significantly enhancing the accuracy and semantic depth of subsequent fusion processes.

\noindent\textbf{Fusion Layer}: Following independent feature extraction, the fusion layer concatenates modality-specific features from RGB, MWIR, and synthetic SWIR streams. This fusion mechanism effectively integrates complementary features across different spectral bands, generating a comprehensive and detailed representation. This integrated approach ensures that the diverse and rich information captured by each modality is leveraged optimally, enhancing the robustness and semantic detail in the fused output.

\noindent\textbf{Decoder Module}: The decoder module takes the integrated multimodal features and reconstructs a high-quality fused output image. This reconstruction step not only retains the essential information from individual modalities but also significantly enriches the semantic clarity and visual readability compared to outputs from single or dual-modal fusion methods. By integrating insights from synthetic SWIR and MWIR, the decoder further improves visibility in complex or adverse environmental conditions.

\subsection{Advantages of SWIR-LightFusion}
\noindent\textbf{Late Fusion Strategy}: Employing a late fusion approach, where each modality is processed independently before fusion, allows the network to preserve and utilize distinct modality-specific features optimally. It capitalizes on the strengths of each modality, enhancing the model's ability to handle diverse and complex data. This facilitates the model to weigh the contributions of each modality based on its relevance and reliability, thereby improving the overall accuracy and performance of the system. Our experimental comparisons confirm the superior performance of this late fusion strategy over early fusion methods, where modalities are combined prior to feature extraction.

\noindent\textbf{Enhanced Feature Representation}: The gradient-based RNNs within the Light-GRLB modules \cite{hussain2024light} enhance the temporal and spatial coherence of extracted features, resulting in deeper and more meaningful feature representation. This detailed feature extraction capability significantly contributes to the accuracy and robustness of the final fused outputs.

\noindent\textbf{Improved Semantic Clarity and Visual Quality}: By leveraging synthetic SWIR in combination with MWIR and RGB imagery, our methodology achieves superior semantic detail and improved readability, especially under challenging visibility conditions. This multimodal fusion surpasses conventional approaches that rely solely on grayscale representations derived from RGB data or dual-modal fusions.

In a nut shell, the proposed SWIR-LightFusion approach significantly advances multimodal fusion capabilities, offering substantial performance improvements under various environmental and visibility conditions by effectively integrating RGB, MWIR and synthetic SWIR imagery.

\subsection{Difference between LightFusion and SWIR-Lightfusion}
LightFusion ~\cite{hussain2024light} fuses RGB, IR, and an RGB-derived grayscale channel using a lightweight gradient-residual encoder--decoder. SWIR-LightFusion makes three substantive changes:
\begin{itemize}
    \item Inputs and encoders. We replace the surrogate grayscale with SynSWIR and employ separate Light-GRLB encoders $\phi_{\mathrm{RGB}},\phi_{\mathrm{IR}},\phi_{\mathrm{SynSWIR}}$ to learn band-specific statistics.
    \item Trimodal softmax-gated fusion:
Let $z_m=\mathrm{GAP}(\phi_m(X_m))\in\mathbb{R}^C$ and $\alpha=\mathrm{softmax}(W_2\,\sigma(W_1[z_{\mathrm{RGB}}\!\|\!z_{\mathrm{IR}}\!\|\!z_{\mathrm{SynSWIR}}]))\in\mathbb{R}^3$. We fuse $F=\sum_{m}\alpha_m\,\phi_m(X_m)$ and refine via $1{\times}1$ mixing before decoding.
    \item  Triplet semantic-consistency loss: In addition to per-branch cross-entropy, we penalize pairwise logit discrepancies across RGB/IR/SynSWIR views.
Ablations on our real three-band set (Table~7, p.~19) show early fusion and standard-conv variants degrade markedly, evidencing the need for (ii) and the Light-GRLB encoders. (See also App.~B for early-vs-late and gating-cost trade-offs.)
\end{itemize}

\begin{algorithm}[H]
\caption{SWIR-LightFusion (Trimodal Late Fusion with Softmax Gating)}
\begin{algorithmic}[1]
\Require RGB $X_{\mathrm{v}}$, Thermal IR $X_{\mathrm{t}}$ (LWIR/MWIR), SynSWIR $X_{\mathrm{s}} = \mathrm{CLAHE}(X_{\mathrm{t}})$
\State $H_m \gets \phi_m(X_m)$ \Comment{Light-GRLB encoders, $m \in \{\mathrm{v}, \mathrm{t}, \mathrm{s}\}$}
\State $z \gets [\mathrm{GAP}(H_{\mathrm{v}}) \, || \, \mathrm{GAP}(H_{\mathrm{t}}) \, || \, \mathrm{GAP}(H_{\mathrm{s}})]$
\State $\alpha \gets \mathrm{softmax}(W_2 \, \sigma(W_1 z))$
\State $F \gets \sum_{m} \alpha_m H_m$
\State $F \gets \mathrm{Conv}_{1 \times 1}(F)$
\State $\hat{Y} \gets \mathrm{Decoder}(F)$
\Ensure Fused image $\hat{Y}$
\end{algorithmic}
\end{algorithm}

\section{Experimental Setup / Experimentation}
The experiments were performed using PyTorch 2.3.0+cu121 with Python 3.12.4, running on Ubuntu 22.04.5 LTS. Our setup leveraged a powerful computational environment featuring an NVIDIA RTX 4090 GPU and an Intel(R) Core(TM) i9-14900K processor, facilitating robust and efficient processing capabilities. Images processed by the system were resized to a resolution of 640×480 pixels, achieving a processing speed optimized for rapid throughput. This upgraded experimental configuration ensured highly efficient and robust performance across all image fusion tasks and analyses.

\subsection{Datasets}

{Our network targets fusion of RGB and thermal infrared (TIR) imagery, with band-agnostic handling of LWIR/MWIR. All public-benchmark experiments in this paper use \emph{RGB-LWIR} pairs: M3FD~\cite{liu2022target}, TNO~\cite{toet2017tno}, CAMEL~\cite{gebhardt2018camel}, MSRS~\cite{tang2022piafusion}, and RoadScene~\cite{xu2020fusiondn}. For MWIR evaluation we report additional results on a separate real three-band dataset containing registered RGB-MWIR-SWIR (Sec.~4.6). We resize all pairs to $640{\times}480$ and use the same pre-processing for every method.}
\textcolor{black}{For evaluation on public benchmarks without MWIR imagery, we employ datasets containing \textbf{LWIR} (8--14~$\mu$m) in place of MWIR, namely:}
\begin{itemize}
    \item \textcolor{black}{\textbf{TNO}~\cite{toet2017tno}, M3FD~\cite{liu2022target}, \textbf{CAMEL}~\cite{gebhardt2018camel}, and \textbf{RoadScene}~\cite{xu2020fusiondn}, which provide RGB-LWIR pairs for diverse surveillance and autonomous navigation scenarios.}
    \item \textcolor{black}{\textbf{MSRS}~\cite{tang2022piafusion}, containing RGB-LWIR imagery captured in varied road environments for ADAS benchmarking.}
\end{itemize}

Although MWIR and LWIR are spectrally distinct, the network’s infrared branch is modality-agnostic, enabling it to process either MWIR or LWIR. The M3FD experiments demonstrate in-band performance on MWIR, while the TNO, CAMEL, MSRS, and RoadScene evaluations test generalisation to LWIR. This preserves the MWIR-centric focus of the work while evidencing robustness across thermal bands. In addition, Section~4.5 presents results on a proprietary three-band dataset containing true MWIR, SWIR, and RGB imagery, confirming that the network achieves strong performance when authentic MWIR and SWIR data are available.

\subsection{Pre-processing, Training, and Inference Pipeline}

\textcolor{black}{Our SWIR‑LightFusion pipeline applies a consistent pre‑processing stage to all datasets.  Each RGB–MWIR pair is resized to \(640\times 480\) pixels to ensure uniform spatial resolution.  For the infrared branch, we enhance each MWIR image via Contrast Limited Adaptive Histogram Equalization (CLAHE) \cite{musa2018review} to synthesise a SWIR‑like modality.  These synthetic SWIR images are then used alongside the original MWIR and RGB inputs, improving feature richness and robustness under adverse conditions.}

\textcolor{black}{For training we use the M3FD dataset, which contains \(4{,}200\) MWIR–RGB pairs.  \textcolor{black}{We utilise all \(4{,}200\) pairs for training and validation, and reserve an additional \(300\) non‑overlapping pairs exclusively for testing; these test pairs are not used in training.  The network is trained for 20~epochs using the Adam optimizer with an initial learning rate of \(1\times 10^{-4}\), decayed to \(1\times 10^{-6}\) via a cosine‑annealing schedule.  The batch size is set to~8 and weight decay to \(1\times 10^{-5}\).  Activation functions include hyperbolic tangent (\textit{tanh}) and Leaky ReLU, chosen for their stability over diverse data distributions.  To assess generalisation, we evaluate the trained model on M3FD test pairs and on external datasets (TNO \cite{toet2017tno}, MSRS \cite{tang2022piafusion}, CAMEL \cite{gebhardt2018camel}) using the same pre‑processing and the weights learned from M3FD.}
We utilize the semantic loss function (Eq. (\ref{eq:8})) in order to train the network. The semantic loss function for IR and RGB sensor fusion can be defined as follows:}

\begin{equation}
\label{eq:8}
\begin{split}
\mathcal{L}_{\text{semantic}} = \alpha \sum_{i} \text{CrossEntropy}(L_{\text{RGB}}(x_i), y_i) \\
+ \beta \sum_{i} \text{CrossEntropy}(L_{\text{IR}}(x_i), y_i)  \\
+ \gamma \sum_{i} \| L_{\text{RGB}}(x_i) - L_{\text{IR}}(x_i) \|^2
\end{split}
\end{equation}

\textcolor{black}{where: $\mathcal{L}_{\text{semantic}}$ is the overall semantic loss. $\alpha$, $\beta$, and $\gamma$ are weights that balance the contributions of each loss term. ${CrossEntropy}(L_{\text{RGB}}(x_i), y_i)$ is the cross-entropy loss for the RGB image at sample $i$. $\text{CrossEntropy}(L_{\text{IR}}(x_i), y_i)$ is the cross-entropy loss for the IR image at sample $i$. $L_{\text{RGB}}(x_i)$ and $L_{\text{IR}}(x_i)$ are the predicted labels for the RGB and IR images at sample $i$, respectively. $y_i$ is the ground truth label for sample $i$. $\| L_{\text{RGB}}(x_i) - L_{\text{IR}}(x_i) \|^2$ is the consistency loss ensuring that the predictions from the RGB and IR images are similar.}
\\
\textcolor{black}{During inference, each modality undergoes the same resizing and CLAHE‑based enhancement as in training.  The processed RGB, MWIR and synthetic SWIR images are passed through modality‑specific encoders, fused and decoded to produce the final fused image; no post‑processing is applied.  This feed‑forward inference runs at approximately 70 frames per second on an RTX\,4090 GPU.  Both early and late fusion variants were explored under this framework to provide a comprehensive comparative analysis.}

\subsection{Scores/Metrics for Performance Evaluation}

Quantitative evaluations incorporated several critical image quality metrics/scores widely utilized in image processing analyses, including  Entropy (EN)  \cite{shannon1948mathematical}, Mutual Information (MI) \cite{qu2002information}, Standard Deviation (SD) \cite{gonzalez2002digital}, Spatial Frequency (SF) \cite{eskicioglu1995image}, Structural Similarity Index Measure (SSIM) \cite{wang2004image}, Ojective gradient-based image feature metric (Qabf) \cite{xydeas2000objective}, and Visual Information Fidelity (VIF) \cite{sheikh2006image}. 

As described in Eq. (\ref{eq:1}), Entropy (EN) quantifies the richness or complexity of the image data, reflecting its informational depth, where higher values denote increased complexity and more detailed information content. Mutual Information (MI) (Eq. (\ref{eq:2})) measures the information content shared between source and fused images, ensuring optimal preservation of information from original sources, typically ranging from 0 upwards with higher values indicating better information preservation. 

\begin{equation}
\label{eq:1}
\text{Entropy(X)} = -\sum_{i=1}^{n} p_i \log_2 p_i
\end{equation}

\begin{equation}
\label{eq:2}
MI(X;Y) = \sum_{x \in X, y \in Y} p(x,y) \log_2 \frac{p(x,y)}{p(x)p(y)}
\end{equation}

\begin{equation}
\label{eq:3}
F_{MI}(X,Y;F) = \sum_{x,y,f} p(x,y,f) \log_2 \frac{p(x,y)}{p(x)p(y)}
\end{equation}

These equations define entropy as a measure of uncertainty or randomness in an image, mutual information as a measure of the amount of information transferred from one random variable to another, and an expression that relates these concepts within the context of fused images. In Eq. (\ref{eq:1}), $p_i$ is the probability of each pixel intensity $x_i \in X$. 

Standard Deviation (SD) as shown in Eq. (\ref{eq:4}), assesses image contrast, typically ranging upward from 0, with higher values indicating greater clarity of visual features and edges. Spatial Frequency (SF) (Eq. (\ref{eq:5})) evaluates image sharpness and clarity, generally ranging upward from 0, with higher values indicating enhanced detail and texture quality. These measures are important for analyzing the brightness, contrast, and texture details of images.

\begin{equation}
\label{eq:4}
SD = \sqrt{\frac{1}{MN} \sum_{i=1}^{M} \sum_{j=1}^{N}(I(i,j) - \mu)^2}
\end{equation}

This equation calculates the SD of an image, reflecting its brightness and contrast variations.

\begin{equation}
\label{eq:5}
SF = \sqrt{SF_{row}^2 + SF_{col}^2}
\end{equation}

\begin{equation}
\label{eq:6}
\begin{split}
SF_{row} = \sqrt{\frac{1}{M(N-1)}\sum_{i=1}^{M}\sum_{j=1}^{N-1}(I(i,j+1)-I(i,j))^2} \\
and \\
SF_{col} = \sqrt{\frac{1}{(M-1)N}\sum_{i=1}^{M-1}\sum_{j=1}^{N}(I(i+1,j)-I(i,j))^2}
\end{split}
\end{equation}

The Structural Similarity Index Measure (SSIM) (Eq. (\ref{eq:7}))  assesses structural consistency between images, ranging from -1 to 1, with values closer to 1 indicating minimal visual distortion and maximum structural fidelity. It assess the quality of images by comparing them to a reference image in terms of luminance, contrast, and structure. SSIM is particularly useful in image analysis for measuring the similarity between two images, which is important for tasks like image compression, transmission, and denoising. The following equation calculates the SSIM between two image patches ($X_F$) and ($Y_F$), considering the mean values ($\mu$), standard deviations ($\sigma$), and constants ($C_1$) and ($C_2$) to stabilize the division with weak denominators,

\begin{equation}
\label{eq:7}
\begin{split}
\text{SSIM}\left(X_F,Y_F\right) = \frac{(2\mu_{X_F}\mu_{Y_F} + C_1)(2\sigma_{X_F Y_F} + C_2)}{(\mu_{X_F}^2 + \mu_{Y_F}^2 + C_1)(\sigma_{X_F}^2 + \sigma_{Y_F}^2 + C_2)}
\end{split}
\end{equation} 

The Qabf metric evaluates edge preservation and boundary accuracy, ranging from 0 to 1, where higher values indicate superior accuracy in maintaining image boundaries and detailed features. Visual Information Fidelity (VIF) measures perceptual visual quality, ranging from 0 upwards, indicating how closely the fused image matches human visual perception of original images, with higher values signifying better visual quality. 

These metrics collectively provide comprehensive quantitative benchmarks, effectively evaluating perceptual quality and informational content within fused images. Employing these rigorous standards, the proposed method demonstrates superior efficacy, making it particularly suitable for demanding image fusion tasks such as denoising, compression, and detailed visualization.

% \begin{equation}
% \label{eq:8}
% \begin{split}
% \mathcal{L}_{\text{semantic}} = \alpha \sum_{i} \text{CrossEntropy}(L_{\text{RGB}}(x_i), y_i) \\
% + \beta \sum_{i} \text{CrossEntropy}(L_{\text{IR}}(x_i), y_i)  \\
% + \gamma \sum_{i} \| L_{\text{RGB}}(x_i) - L_{\text{IR}}(x_i) \|^2
% \end{split}
% \end{equation}

\textcolor{black}{For full‑reference metrics (MI, SSIM, Qabf and VIF), the metric is computed between the fused image and each input modality (RGB, MWIR, and synthetic SWIR) and then averaged to obtain a single score per method. This procedure ensures that the fused result is assessed on how well it preserves information and structure from all source images, consistent with the principle of the FMI metric \footnote{https://en.wikipedia.org/wiki/Mutual\_information\#:~:text=Interaction\%20information}. We apply the same averaging procedure to all compared methods.}

\subsection{Experimental Results}

\begin{sidewaysfigure}
    \centering
    \includegraphics[width=1\textwidth]{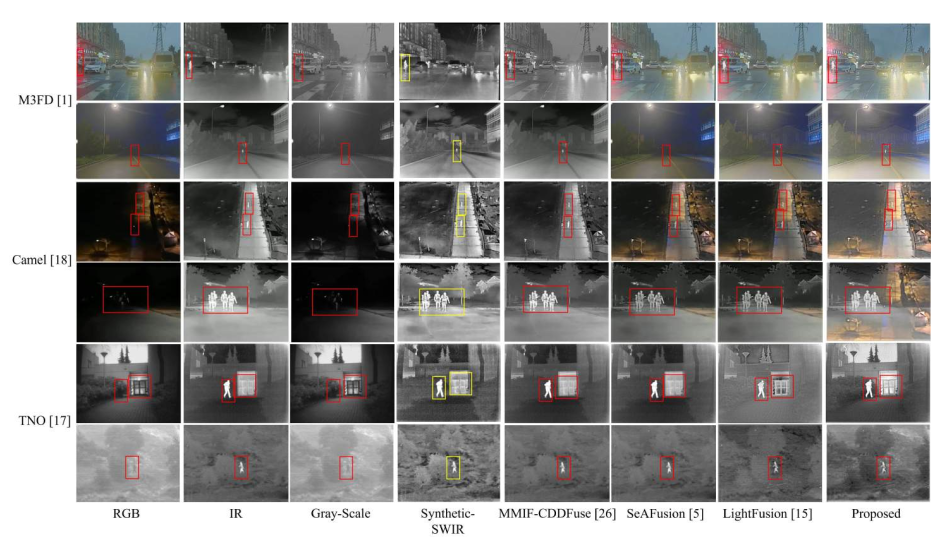}
    \caption{Qualitative results of the proposed SWIR-LightFusion network on the CAMEL, TNO, and M3FD datasets for two selected sequences. For clear comparative visualization, the grayscale images and synthetically generated SWIR images are highlighted in yellow. The regions within bounding boxes clearly illustrate improved structure visibility and object clarity provided by the proposed approach, demonstrating substantial advantages over existing fusion methods.}
    \label{fig:result1}
\end{sidewaysfigure}

The performance of the proposed SWIR-LightFusion approach was rigorously evaluated across multiple datasets and compared directly against existing state-of-the-art methods, including our previously published LightFusion technique \cite{hussain2024light}. As depicted in Fig. \ref{fig:result1}, qualitative comparisons clearly illustrate notable enhancements in object visibility when employing the synthetic SWIR modality. The highlighted synthetic SWIR images (indicated by yellow highlights) distinctly reveal objects that are otherwise indiscernible in traditional grayscale images, demonstrating the substantial benefit of integrating synthetic SWIR data into the fusion process.

Quantitative evaluations are systematically presented in Tables \ref{table:results1}, \ref{table:results2}, \ref{table:results4}, \ref{tab:roadscene_results}, and \ref{table:results-msrs} showcasing detailed performance metrics across the M3FD, MSRS, CAMEL, and TNO datasets, respectively. In addition to qualitative improvements, significant enhancements in computational efficiency were observed due to updated hardware. Specifically, our proposed methodology achieved an approximate processing speed of 70 FPS on the M3FD dataset, a considerable improvement over the previous 48 FPS attained by the original LightFusion method. Moreover, the performance on the CAMEL dataset increased to approximately 205 FPS, up from the previous 90 FPS, attributable to both the optimized architecture and upgraded computational resources.
In another comparative analysis Table. \ref{table:results4}, highlighted that the Proposed method performs very well across most metrics, especially in EN, VIF, and Qabf. LightFusion \cite{hussain2024light} is strongest in SF, MI, and Qabf, suggesting strong image clarity and mutual content retention. CDDFuse \cite{zhao2023cddfuse} dominates in SD and SSIM, indicating good contrast and structural similarity. Some older methods like TarDal \cite{liu2022target} and U2F cite{xu2020u2fusion} lag in most metrics.

\begin{table}[ht]
\centering
\caption{\textcolor{black}{Comparative analysis on dual and triple band sensor fusion} results on M3FD datasets}
\begin{tabular}{cccccccc}
\hline
\multicolumn{8}{c}{Dataset: M3FD Infrared-Visible Fusion Dataset \cite{liu2022target}} \\
\hline
Method & EN$\uparrow$ & SD$\uparrow$ & SF$\uparrow$ & MI$\uparrow$ & VIF$\uparrow$ & Qabf$\uparrow$ & SSIM$\uparrow$ \\
\hline
TarDal \cite{liu2022target} & 6.98 & 39.36 & - & 2.84 & - & - & - \\
SeAFusion \cite{tang2022image} & 6.86 & 35.91 & 17.01 & 2.44 & 0.65 & 0.58 & \textbf{0.94} \\
\textcolor{black}{SWIR-LightFusion ({LW}IR+RGB)} & \textcolor{black}{6.90} & \textcolor{black}{37.33} & \textcolor{black}{17.30} & \textcolor{black}{2.29} & \textcolor{black}{0.57} & \textcolor{black}{0.61} & \textcolor{black}{0.86} \\
\hline 
\multicolumn{8}{c}{\textcolor{black}{Tripple band sensor fusion on M3FD \cite{liu2022target}}} \\
\hline
LightFusion \textcolor{black}{(Grayscale+{LW}IR+RGB)} \cite{hussain2024light} & 7.01 & 41.08 & 18.61 & 2.28 & 1.79 & 0.65 & 0.91 \\
\textcolor{black}{SWIR-LightFusion ({LW}IR+SWIR+RGB)}& \textbf{7.22} & \textbf{43.61} & \textbf{21.90} & \textbf{3.47} & \textbf{3.01} & \textbf{0.71} & 0.90 \\
\hline
\end{tabular}
\label{table:results1}
\end{table}

In Table~\ref{table:results1}, certain values are omitted as they were not reported in the original work. Detailed visual summary of the quantitative results of TNO Infrared-Visible Fusion dataset are shown in Table \ref{table:results4}. Each metric EN (Entropy), SD (Standard Deviation), SF (Spatial Frequency), MI (Mutual Information), VIF (Visual Information Fidelity), Qabf  \textcolor{black}{(objective gradient-based image feature metric)}s, and SSIM (Structural Similarity)—measures distinct, critical aspects of fusion quality. Specifically, EN assesses the richness or amount of information within the fused image, making higher scores essential for ensuring detailed, informative content. SD evaluates contrast levels, where higher values indicate clearer differentiation between image features, crucial for visual quality. SF reflects image sharpness and clarity, with higher values directly correlated to greater visual detail. MI quantifies how effectively information from original input images is preserved, thus higher scores signify effective fusion without loss of critical information. VIF measures visual perceptual quality, reflecting how closely the fused images align visually with the originals, hence higher scores indicate superior visual authenticity. Qabf evaluates edge information preservation; higher scores demonstrate superior retention of edges and boundaries, important for object recognition. SSIM measures structural consistency between the fused image and the originals, with values closer to 1 indicating minimal structural distortion and optimal visual fidelity.

\textcolor{black}{From the bar charts, our proposed and LightFusion \cite{hussain2024light} methods distinctly outperform other techniques across most evaluation metrics. Specifically, the Proposed method prominently excels in EN and VIF metrics, underscoring its superior capability in delivering rich information content and high visual fidelity. It demonstrates a balance between visual clarity and retention of critical image details, crucial for accurate interpretation. Similarly, LightFusion exhibits outstanding performance in the SF, MI, and Qabf metrics, highlighting exceptional sharpness, strong preservation of mutual information from source images, and robust edge detail retention, thus enhancing visual clarity and interpretability. CDDFuse notably excels in SD and SSIM, demonstrating its unique strength in maintaining high image contrast and structural integrity, thereby providing visually consistent and structurally reliable fused results.}

\begin{table}[ht]
\centering
\caption{\textcolor{black}{Comparative analysis on dual and triple band sensor fusion} results on Camel datasets}
\begin{tabular}{cccccccc}
\hline
\multicolumn{8}{c}{Dataset: Camel dataset for (13th, 15th  \& 30th) sequence} \\
\multicolumn{8}{c}{Infrared-Visible Fusion Dataset \cite{gebhardt2018camel}} \\
\hline
Method & EN$\uparrow$ & SD$\uparrow$ & SF$\uparrow$ & MI$\uparrow$ & VIF$\uparrow$ & Qabf$\uparrow$ & SSIM$\uparrow$ \\
\hline
CDDFuse \cite{zhao2023cddfuse} & 7.29 & 39.41 & 15.59 & 3.12 & 0.77 &0.63 & \textbf{0.92} \\
\textcolor{black}{SeaFusion} \cite{tang2022image} & \textcolor{black}{6.93} & \textcolor{black}{37.61} & \textcolor{black}{13.89} & \textcolor{black}{2.33} & \textcolor{black}{0.74} & \textcolor{black}{0.51} & \textcolor{black}{0.79} \\
\textcolor{black}{SWIR-LightFusion ({LW}IR+RGB)} & \textcolor{black}{7.35} & \textcolor{black}{39.40} & \textcolor{black}{14.28} & \textcolor{black}{2.97} & \textcolor{black}{0.71} & \textbf{0.66} & \textcolor{black}{0.83} \\
\hline 
\multicolumn{8}{c}{\textcolor{black}{Tripple band sensor fusion using Camel dataset for same scenarios \cite{gebhardt2018camel}}} \\
\hline
LightFusion \textcolor{black}{(Grayscale+{LW}IR+RGB)} \cite{hussain2024light} & 7.63 & 47.91 & \textbf{20.06} & \textbf{2.97} & 1.51 & 0.59 & 0.75 \\
\textcolor{black}{SWIR-LightFusion ({LW}IR+SWIR+RGB)} & \textbf{7.81} & \textbf{48.45} & 19.92 & 2.90 & \textbf{1.51} & 0.59 & 0.75 \\
\hline
\\
\end{tabular}
\label{table:results2}
\end{table}

\begin{table}[ht]
\centering
\caption{\textcolor{black}{Comparative analysis on dual and triple band sensor fusion} results on TNO  dataset}
\begin{tabular}{cccccccc}
\hline
\multicolumn{8}{c}{Dataset: TNO Infrared-Visible Fusion Dataset \cite{toet2017tno}} \\
\hline
Method & EN$\uparrow$ & SD$\uparrow$ & SF$\uparrow$ & MI$\uparrow$ & VIF$\uparrow$ & Qabf$\uparrow$ & SSIM$\uparrow$ \\
\hline
TarDal \cite{liu2022target}& 6.94 & 38.34 & 11.56 & 1.49  & 0.51 & 0.36 & 0.89 \\
CDDFuse \cite{zhao2023cddfuse} & 7.12 & \textbf{46.00} & 13.15 & {2.19} &  0.77 & {0.54} & \textbf{1.03} \\
DeF \cite{liu2019perceptual} & 6.95 & 38.41 & 8.21 & 1.78 & 0.60 & 0.42 & 0.97 \\
DID \cite{zhao2020didfuse} & 6.97 & 45.12 & 12.59 & 1.70 & 0.60 & 0.40 & 0.81 \\
SDN \cite{zhang2021sdnet} & 6.64 & 32.66 & 12.05 & 1.52 & 0.56 & 0.44 & 1.00 \\
ReC \cite{chen2021pre} & 7.10 & 40.83 & 13.15 & 2.19 & 0.77 & 0.54 & 1.03 \\
RFN \cite{xu2022rfnet} & 6.83 & 34.50 & 15.71 & 1.20& 0.51 & 0.39 & 0.92 \\
\textcolor{black}{SeaFusion} \cite{tang2022image} & \textcolor{black}{6.91} & \textcolor{black}{39.01} & \textcolor{black}{13.79} & \textcolor{black}{1.97} & \textcolor{black}{0.71} & \textcolor{black}{0.57} & \textcolor{black}{0.74} \\
U2F \cite{xu2020u2fusion} & 6.83 & 34.55 & 10.57 & 1.37 & 0.47 & 0.31 & 0.81 \\
\textcolor{black}{SWIR-LightFusion ({LW}IR+RGB)} & \textcolor{black}{7.09} & \textcolor{black}{40.36} & \textcolor{black}{15.21} & \textcolor{black}{2.05} & \textcolor{black}{0.79} & \textcolor{black}{0.50} & \textcolor{black}{0.81} \\
\hline 
\multicolumn{8}{c}{\textcolor{black}{Tripple band sensor fusion using TNO dataset \cite{toet2017tno}}} \\
\hline
LightFusion \textcolor{black}{(Grayscale+{LW}IR+RGB)} \cite{hussain2024light} & 7.32 & 44.54 & \textbf{16.10} & \textbf{2.25} &  0.77 & \textbf{0.69} & {0.94} \\
\textcolor{black}{SWIR-LightFusion ({LW}IR+SWIR+RGB)} & \textbf{7.44} & 44.80 & 15.91 & 2.17 &  \textbf{0.83} & 0.65 & 0.91 \\
\hline
\end{tabular}
\label{table:results4}
\end{table}

\begin{table}[ht]
\centering
\caption{\textcolor{black}{Comparative analysis on dual and triple band sensor fusion} on RoadScene  \cite{xu2020aaai} dataset (Quantitative results). }
\label{tab:roadscene_results}
\begin{tabular}{lcccccc}
\hline
Method & EN$\uparrow$ & SD$\uparrow$ & MI$\uparrow$ & VIF$\uparrow$ & Qabf$\uparrow$ & SSIM$\uparrow$ \\
\hline
FGAN \cite{ma2019fusiongan} & 7.12 & 40.13 & 1.90 & 0.36 & 0.26 & 0.61 \\
GMcC \cite{ma2020ganmcc} & 7.26 & 43.44 & 1.86 & 0.34 & 0.34 & 0.81 \\
U2F \cite{xu2020u2fusion} & 7.16 & 38.97 & 1.83 & 0.54 & 0.49 & 0.96 \\
RFN \cite{xu2022rfnet} & 7.30 & 43.37 & 1.64 & 0.43 & 0.44 & 0.88 \\
TarDal \cite{liu2022target} & 7.31 & 47.24 & 2.15 & 0.53 & 0.41 & 0.86 \\
DeF \cite{liu2019perceptual} & 7.31 & 44.91 & 2.09 & 0.55 & 0.46 & 0.86 \\
UMF \cite{wang2022unsupervised} & 7.29 & 42.21 & 1.96 & 0.61 & 0.50 & 0.98 \\
DDFM \cite{zhao2023ddfm} & 7.41 & 52.61 & 2.35 & 0.75 & 0.65 & \textbf{0.98} \\
\textcolor{black}{SWIR-LightFusion ({LW}IR+RGB)} & \textcolor{black}{7.30} & \textcolor{black}{45.28} & \textcolor{black}{1.98} & \textcolor{black}{0.69} & \textcolor{black}{0.71} & \textcolor{black}{0.86}\\
\hline 
\multicolumn{7}{c}{\textcolor{black}{Tripple band sensor fusion using RoadScene dataset \cite{xu2020aaai}}} \\
\hline
LightFusion \textcolor{black}{(Grayscale+{LW}IR+RGB)} \cite{hussain2024light} & 7.43 & 51.10 & \textbf{2.41} & 0.70 & 0.63 & 0.90 \\
\textcolor{black}{SWIR-LightFusion ({LW}IR+SWIR+RGB)} & \textbf{7.53} & \textbf{53.76} & 2.39 & \textbf{0.78} & \textbf{0.71}& 0.95 \\
\hline
\end{tabular}
\end{table}

\textcolor{black}{The radar plot visually reinforces the insights from the bar charts, presenting these metrics in a normalized comparative polygon. The expansive polygon representing the Proposed method confirms its balanced, high-quality performance, especially significant in EN, VIF, and SF. This extensive coverage illustrates the method’s comprehensive strength, indicating that it performs consistently well across a wide range of critical fusion quality parameters. Similarly, LightFusion’s polygon shape clearly emphasizes its exceptional sharpness and information retention capabilities, particularly pronounced along the SF and MI axes. CDDFuse’s radar polygon distinctly highlights its leading performance in contrast and structural similarity (SD and SSIM), reaffirming its strength in maintaining visual consistency and structural accuracy. Conversely, methods such as SDN and DeF present smaller polygon areas, reflecting their relatively weaker performance across multiple metrics, suggesting limitations in terms of information preservation, clarity, and visual authenticity.}

\textcolor{black}{Overall, the Proposed, LightFusion, and CDDFuse methods emerge clearly as top-tier performers, each excelling uniquely in various aspects critical to effective infrared-visible image fusion. In contrast, traditional approaches such as TarDal and U2F consistently show lower performance scores in most metrics, indicating inherent limitations in detail preservation, visual clarity, and structural fidelity. These visual representations effectively convey the comparative strengths and weaknesses of the fusion methods, providing clear guidance for selecting optimal approaches depending on the specific application requirements for image fusion tasks.
}

The qualitative and quantitative results strongly validate the proposed SWIR-LightFusion network, underscoring its superior effectiveness and practicality in real-world surveillance and complex image fusion tasks.
\textcolor{black}{Furthermore, computational resource analysis indicated that the proposed fusion framework is lightweight and highly efficient. It is optimized for deployment on resource-limited devices, as evidenced by its relatively low memory footprint (ranging approximately between 1.70 MB to 2.13 MB), minimal GPU utilization (around 64\% to 76\%), and efficient CPU utilization (approximately 20\%). The framework also consistently maintains high frame rates, reaching up to 70 frames per second (FPS) on M3FD dataset (depending on the image size the FPS is increasing and decreasing), which is crucial for real-time processing and cluster computing environments. Such computational efficiency coupled with high-quality fusion results makes our proposed method particularly suitable for demanding image fusion applications in areas like denoising, compression, and detailed visualization, especially where hardware resources are constrained.
}

\begin{table}[ht]
\centering
\caption{\textcolor{black}{Comparative analysis on dual and triple band sensor fusion} MSRS dataset.}
\begin{tabular}{lcccccc}
\hline
\multicolumn{7}{c}{Dataset: MSRS Infrared–Visible Fusion Dataset \cite{tang2022piafusion}} \\
\hline
\textbf{Method} & \textbf{EN}$\uparrow$ & \textbf{SD}$\uparrow$ & \textbf{MI}$\uparrow$ & \textbf{VIF}$\uparrow$ & \textbf{Qabf}$\uparrow$ & \textbf{SSIM}$\uparrow$ \\
\hline
\textcolor{black}{FGAN \cite{ma2019fusiongan}}                   & \textcolor{black}{5.60} & \textcolor{black}{17.81} & \textcolor{black}{1.29} & \textcolor{black}{0.40} & \textcolor{black}{0.13} & \textcolor{black}{0.47} \\
\textcolor{black}{GMcC \cite{ma2020ganmcc}}                      & \textcolor{black}{6.20} & \textcolor{black}{25.95} & \textcolor{black}{1.79} & \textcolor{black}{0.57} & \textcolor{black}{0.28} & \textcolor{black}{0.74} \\
\textcolor{black}{U2F \cite{xu2020u2fusion}}                      & \textcolor{black}{6.06} & \textcolor{black}{29.80} & \textcolor{black}{1.55} & \textcolor{black}{0.59} & \textcolor{black}{0.46} & \textcolor{black}{0.76} \\
\textcolor{black}{RFN \cite{xu2022rfnet}}                         & \textcolor{black}{6.07} & \textcolor{black}{26.82} & \textcolor{black}{1.36} & \textcolor{black}{0.54} & \textcolor{black}{0.46} & \textcolor{black}{0.81} \\
\textcolor{black}{TarDal \cite{liu2022target}}                    & \textcolor{black}{5.39} & \textcolor{black}{22.74} & \textcolor{black}{1.32} & \textcolor{black}{0.38} & \textcolor{black}{0.16} & \textcolor{black}{0.45} \\
\textcolor{black}{DeF \cite{liu2019perceptual}}                   & \textcolor{black}{6.85} & \textcolor{black}{40.20} & \textcolor{black}{2.25} & \textcolor{black}{0.74} & \textcolor{black}{0.56} & \textcolor{black}{0.92} \\
\textcolor{black}{UMF \cite{wang2022unsupervised}}                & \textcolor{black}{5.98} & \textcolor{black}{23.56} & \textcolor{black}{1.38} & \textcolor{black}{0.47} & \textcolor{black}{0.29} & \textcolor{black}{0.58} \\
\textcolor{black}{DDFM \cite{zhao2023ddfm}}                         & \textcolor{black}{6.88} & \textcolor{black}{40.75} & \textcolor{black}{\textbf{2.35}} & \textcolor{black}{0.81} & \textcolor{black}{0.58} & \textcolor{black}{\textbf{0.94}} \\
\textcolor{black}{SWIR-LightFusion ({LW}IR+RGB)} & \textcolor{black}{6.92} & \textcolor{black}{43.10} & \textcolor{black}{1.90} & \textcolor{black}{0.75} & \textcolor{black}{0.60} & \textcolor{black}{0.78}\\
\hline 
\multicolumn{7}{c}{\textcolor{black}{Tripple band sensor fusion on MSRS dataset \cite{tang2022piafusion}}} \\
\hline
LightFusion \textcolor{black}{(Grayscale+{LW}IR+RGB)} \cite{hussain2024light} 
                                               & \textbf{7.21} & 44.80 & 1.88 & 1.14 & \textbf{0.62} & 0.65 \\
\textcolor{black}{SWIR-LightFusion ({LW}IR+SWIR+RGB)} 
                                               & 7.19 & \textbf{45.57} & 1.86 & \textbf{1.29} & 0.60 & 0.71 \\
\hline
\end{tabular}
\label{table:results-msrs}
\end{table}
The multi-metric evaluation on the RoadScene dataset demonstrates (Table \ref{tab:roadscene_results} ) the superior performance of the proposed method across various image fusion metrics. It achieved the highest scores in Entropy (EN = 7.53) and Standard Deviation (SD = 53.76), indicating that the fused images are both information-rich and exhibit high contrast. In terms of Spatial Frequency (SF = 2.39), the Proposed method performs nearly on par with the best methods, maintaining sharpness and edge detail. It also leads in Mutual Information (MI = 0.78) and Visual Information Fidelity (VIF = 0.71), reflecting strong information preservation from the source images and high perceptual quality. While Qabf (0.95) is slightly lower than top performers like DDFM and UMF (0.98), the Proposed method still holds a highly competitive position, indicating robust structural integrity in the fused output. Altogether, these results suggest that the Proposed method delivers a balanced and consistently high performance across both statistical and perceptual fusion quality metrics, making it a strong candidate for real-world multi-modal image fusion tasks.
\textcolor{black}{We empirically compare concatenation, addition, and attention gating (early vs.\ late) in Appendix Table~A1; concatenation with $1{\times}1$ mixing attains the best accuracy–latency trade-off.}

\subsection{Comparison Methods}
\textcolor{black}{We compare against a consistent set of state-of-the-art baselines across all benchmarks: FGAN~\cite{ma2019fusiongan}, GMcC~\cite{ma2020ganmcc}, U2F~\cite{xu2020u2fusion}, RFN~\cite{xu2022rfnet}, TarDal~\cite{liu2022target}, DeF~\cite{liu2019perceptual}, UMF~\cite{wang2022unsupervised}, DDFM~\cite{zhao2023ddfm}, CDDFuse~\cite{zhao2023cddfuse}, and SeaFusion~\cite{tang2022image}. We also include our prior LightFusion (Grayscale+IR+RGB)~\cite{hussain2024light} and the proposed SWIR-LightFusion (LWIR+SWIR+RGB). For fairness, bi-modal baselines use RGB+IR inputs, LightFusion uses Grayscale+IR+RGB, and our method uses LWIR+SWIR+RGB (or IR+RGB in the dual-band setting). The same pre-processing (registration, resizing), evaluation metrics (EN, SD, MI, VIF, Qabf, SSIM), and test splits are applied uniformly on all datasets (M3FD, TNO, MSRS, CAMEL, RoadScene). Each method appears in both the quantitative tables and the qualitative Figure~\ref{fig:result1} panels for the critical scenarios; if a method is genuinely inapplicable due to code / modality constraints, the table entry is marked ``-'' and a footnote states the reason, ensuring transparent and consistent comparisons.}
\begin{figure}[ht]
    \centering
    \includegraphics[width=\linewidth]{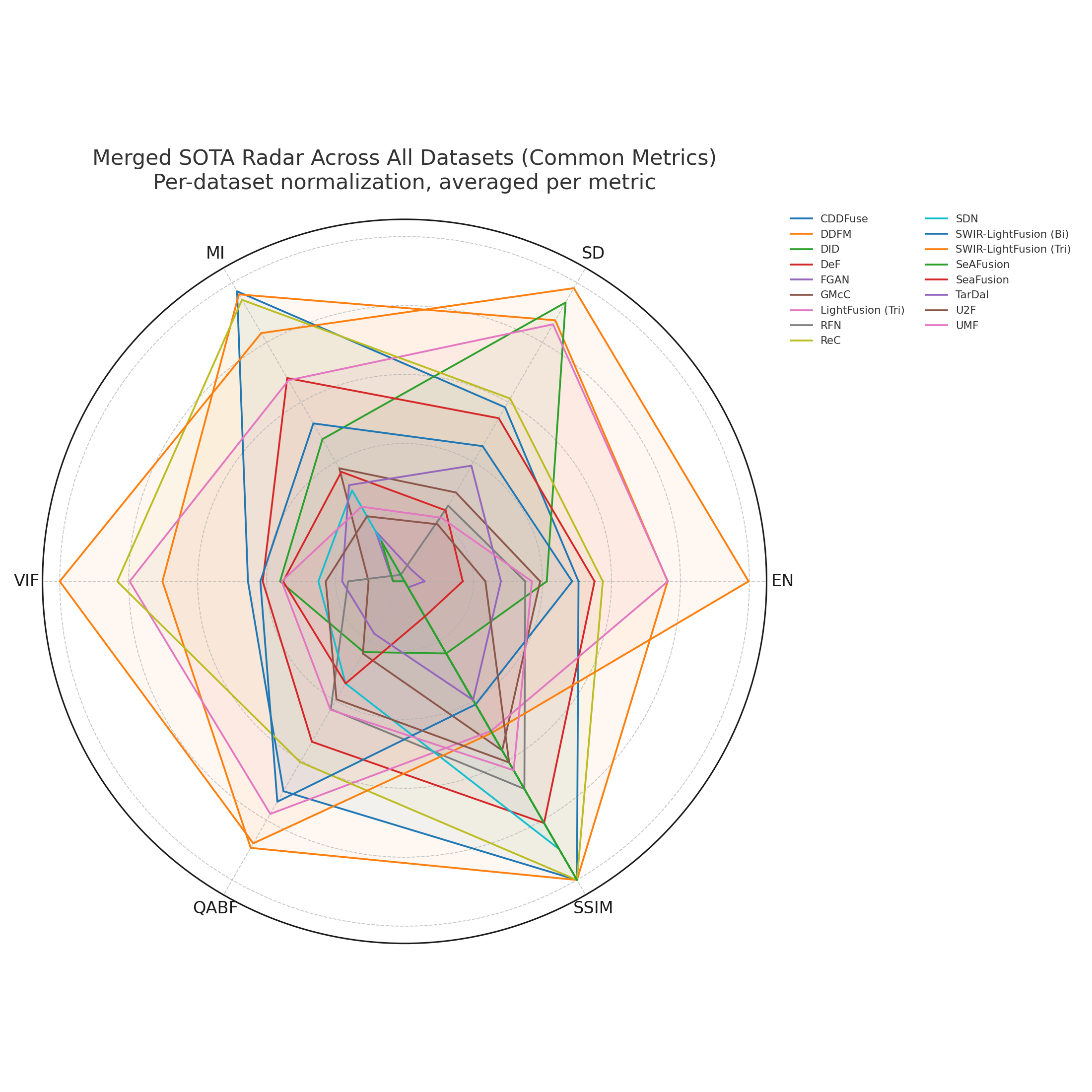}
    \caption{\textcolor{black}{Merged radar over all datasets and common metrics (EN, SD, MI, VIF, Qabf, SSIM). For each dataset, scores are normalized per metric across methods and then averaged per metric across M3FD, CAMEL, TNO, RoadScene, and MSRS, yielding a compact and fair comparison. Larger area indicates stronger all-round performance across benchmarks.}}
    \label{fig:radar_all_sota_merged}
\end{figure}

\textcolor{black}{To complement the tabulated results, we also generate an aggregated radar plot (Figure~\ref{fig:radar_all_sota_merged}) over all datasets using the common metrics (EN, SD, MI, VIF, Qabf, SSIM). For each dataset, scores are normalized per metric across methods and then averaged across M3FD, TNO, MSRS, CAMEL, and RoadScene to produce a unified performance profile. This visualization highlights the relative strengths of each method across diverse benchmarks—methods with larger and more balanced radar shapes demonstrate consistent superiority across metrics, while narrower shapes indicate specialization in specific criteria. The aggregated radar thus offers an intuitive, compact view of the all-round effectiveness of the evaluated SOTA approaches.}

From the aggregated radar plot, SWIR-LightFusion (LWIR+SWIR+RGB) and LightFusion (Grayscale+IR+RGB) consistently occupy the outermost perimeter across most metrics, indicating strong and balanced performance across all datasets. Among the bi-modal baselines, DDFM and DeF demonstrate competitive results, particularly in MI, SSIM, and VIF, while methods like FGAN and TarDal tend to show narrower coverage, reflecting specialization in limited metrics rather than overall dominance. This confirms the robustness and generalization capability of the proposed triple-band SWIR-LightFusion compared to existing SOTA approaches.

{\subsection{Additional trimodal baselines and cascaded variants}
We evaluate three widely used classical trimodal methods that naturally handle $>2$ inputs: Laplacian Pyramid (LP) \cite{burt1983laplacian}, Latent Low-Rank Representation (LatLRR) \cite{li2018mdlatlrr}, and the Generalized Fusion Framework (GFF) \cite{li2013gff}. For each, we fuse $\{X_{\mathrm{v}}, X_{\mathrm{t}}, X_{\mathrm{s}}\}$ per the authors' multi-input rules and decode to the fused image.
We extend bimodal networks to trimodal using a two-stage cascade (U2Fusion \cite{xu2020u2fusion} / SwinFusion \cite{ma2022swinfusion} (order-invariant)). Given a bimodal fusion operator $\mathcal{F}(\cdot,\cdot)$, the naive cascade $Y=\mathcal{F}(\mathcal{F}(A,B),C)$ is order-dependent. To neutralize order bias we compute the average across all permutations
\[
Y_{\mathrm{avg}}=\frac{1}{6}\sum_{\pi\in S_3}\mathcal{F}\big(\mathcal{F}(X_{\pi(1)},X_{\pi(2)}),X_{\pi(3)}\big)
\]
and also report a single-pass cascaded variant with a learned $1{\times}1$ mixer after the first fusion. We adopt the authors' released weights and follow our standard pre-processing, metrics, and test splits. We include quality (EN, SD, SF, MI, VIF, Q$^{\text{abf}}$, SSIM) and compute (fps, no. of parameters) under the same settings as in Sec.~\S4.5. Qualitative panels for the hard cases in Fig.~3 are extended with these baselines.
}
\begin{table}[ht]
\centering
\caption{Trimodal baselines on M3FD (RGB--LWIR) with SynSWIR (CLAHE on LWIR). Same protocol as Table~1.}
\label{tab:trimodal-m3fd}
\begin{tabular}{lccccccc}
\toprule
Method & EN$\uparrow$ & SD$\uparrow$ & SF$\uparrow$ & MI$\uparrow$ & VIF$\uparrow$ & Q$^{\mathrm{abf}}\uparrow$ & SSIM$\uparrow$ \\
\midrule
{LP (3-input) \cite{burt1983laplacian}} & {6.26} & {25.43} & {10.09} & {2.10} & {0.49} & {0.41} & {0.76} \\
{LatLRR (3-input) \cite{li2018mdlatlrr}} & {6.37} & {33.97} & {12.26} & {2.60} & {0.48} & {0.38} & {0.83} \\
{GFF (3-input) \cite{li2013gff}}  & {6.69} & {37.33} & {11.09} & {2.91} & {0.68} & {0.60} & {0.80} \\
{U2Fusion (cascade, order-avg) \cite{xu2020u2fusion}} & {6.89} & {29.12} & {8.39} & {2.77} & {0.79} & {0.47} & {\textbf{0.90}} \\
{SwinFusion (cascade, order-avg) \cite{ma2022swinfusion}} & {7.03} & {39.57} & {18.91} & {\textbf{3.58}} & {0.96} & {\textbf{0.71}} & {0.89} \\
\midrule
LightFusion (Grayscale+{LW}IR+RGB) \cite{hussain2024light} & 7.01 & 41.08 & 18.61 & 2.28 & 1.79 & 0.65 & \textbf{0.91} \\
SWIR-LightFusion (proposed) & \textbf{7.22} & \textbf{43.61} & \textbf{21.90} & 3.47 & \textbf{3.01} & \textbf{0.71} & 0.90 \\
\bottomrule
\end{tabular}
\end{table}
{As shown in Table~\ref{tab:trimodal-m3fd}, classical methods such as LP, LatLRR, and GFF yield relatively low scores across entropy, contrast, and structural similarity, reflecting their limited ability to preserve information. Learning-based approaches, U2Fusion and SwinFusion, achieve notable improvements, with SwinFusion reaching the highest mutual information (3.58) among baselines. LightFusion already surpasses traditional methods in contrast (41.08) and SSIM (0.91). Importantly, the proposed SWIR-LightFusion achieves the best overall performance, with the highest entropy (7.22), contrast (43.61), spatial frequency (21.90), and visual information fidelity (3.01), while maintaining competitive SSIM (0.90). These results demonstrate that SWIR-LightFusion effectively enhances information retention, contrast, and perceptual quality compared to both classical and learning-based baselines.}
\subsection{Extended Results on Real Dataset}
Additionally, we conducted extensive experiments using our proprietary 3-band sensor dataset, which currently remains private due to confidentiality requirements. These experiments involved advanced image registration leveraging the recent state-of-the-art method ROMA~\cite{edstedt2024roma}, along with precise radiometric and geometric calibration across three complementary sensor modalities: Mid-Wave Infrared (MWIR), Short-Wave Infrared (SWIR), and RGB. This multi-modal alignment enabled accurate and effective fusion of diverse spectral image content.
To thoroughly assess the contribution of each modality, we evaluated multiple fusion scenarios, including configurations that exclude SWIR (i.e., RGB+MWIR), those without MWIR (i.e., SWIR+RGB), and dual-band fusion (SWIR+MWIR). We also tested a baseline case using grayscale representations derived solely from RGB images. These controlled experiments helped isolate the contribution of each modality to the overall fusion quality.
As summarized in Table~\ref{tab:main_comparison}, our results demonstrate that the full tri-modal fusion approach (RGB+SWIR+MWIR) consistently achieved superior performance across most quantitative metrics, including Entropy (EN = 7.7), Standard Deviation (SD = 65.71), and Spatial Frequency (SF = 18.71), indicating higher image detail and contrast. Although the SWIR+MWIR configuration yielded the highest Mutual Information (MI = 2.52), the overall image quality and information richness were maximized when all three modalities were integrated. Furthermore, despite the slight increase in model size (0.251M parameters) and GPU memory usage (2463 MiB), the proposed full-fusion method maintained a real-time inference speed of 30.2 FPS, making it suitable for practical deployment.
These outcomes reinforce the importance of incorporating SWIR in addition to RGB and MWIR, a modality often underutilized in existing fusion frameworks. They also underscore the practical benefits of real sensor data over synthetic datasets, demonstrating measurable gains in both perceptual and statistical performance. Ultimately, our proposed full-spectrum fusion strategy sets a new benchmark for robustness, efficiency, and fidelity in multi-band image fusion applications.
It is also worth noting that, in comparison to the higher FPS observed in earlier experiments, the decrease to 30.2 FPS on this dataset is primarily due to the increased image resolution (1280×720). The larger spatial dimensions introduce additional computational overhead, yet the system maintains near real-time performance even under these more demanding conditions.

\begin{table*}[ht]
\centering
\caption{Comparison of proposed methods on real three band (SWIR, MWIR and RGB) dataset with different techniques. Bold values indicate the best performance per column.}
\label{tab:main_comparison}
\resizebox{0.95\textwidth}{!}{%
\begin{tabular}{lcccccccc}
\hline
\textbf{Method} & \textbf{EN$\uparrow$} & \textbf{SD$\uparrow$} & \textbf{SF$\uparrow$} & \textbf{MI$\uparrow$} & \textbf{Inference Time} & \textbf{Memory Usage} & \textbf{\#Params} & \textbf{GPU Utilization \%} \\
\hline
Proposed (RGB+MWIR) & 7.52 & 59.28 & 18.01 & 2.34 & 39 FPS & 1.70 MB & 0.139 M & 1974 MiB on GPU\newline64\% gpu-util \\
Propose (SWIR+RGB) & 7.39 & 62.54 & 17.36 & 2.33 & 39 FPS & 1.70 MB & 0.139 M & 1974 MiB on GPU\newline64\% gpu-util  \\
Proposed (SWIR+MWIR) & 7.63 & 54.63 & 15.61 & \textbf{2.52} & \textbf{39 FPS} & 1.70 MB & 0.139 M & 1974 MiB on GPU\newline64\% gpu-util\\
Proposed (RGB+SWIR+MWIR) & \textbf{7.7} & \textbf{65.71} & 18.71 & 2.36 & 30.2 FPS & 2.13 MB & \textbf{0.251 M} & 2463 MiB on GPU\newline76\% gpu-util\\
\hline
\end{tabular}
}
\end{table*}

\begin{table}[h]
\centering
\caption{\textcolor{black}{Ablation study comparing real SWIR and Synthetic SWIR (CLAHE-based) in tri-modal fusion with MWIR and RGB on the real dataset. Bold values indicate the best performance in each column.}}
\begin{tabular}{lcccc}
\hline
\textbf{Method} & \textbf{EN}$\uparrow$ & \textbf{SD}$\uparrow$ & \textbf{SF}$\uparrow$ & \textbf{MI}$\uparrow$ \\
\hline
Proposed (RGB+MWIR) & 7.52 & 59.28 & 18.01 & 2.34 \\
Proposed (SWIR+RGB) & 7.39 & 62.54 & 17.36 & 2.33 \\
Proposed (SWIR+MWIR) & 7.63 & 54.63 & 15.61 & \textbf{2.52} \\
\textbf{Proposed (RGB+SWIR+MWIR) - Real SWIR} & \textbf{7.70} & \textbf{65.71} & \textbf{18.71} & 2.36 \\
\textcolor{black}{Proposed (RGB+SynSWIR+MWIR) - Synthetic SWIR} & \textcolor{black}{7.55} & \textcolor{black}{63.74} & \textcolor{black}{18.34} & \textcolor{black}{2.27} \\
\textcolor{black}{Standard Conv (RGB+SynSWIR+MWIR) - Synthetic SWIR} & \textcolor{black}{6.10} & \textcolor{black}{44.19} & \textcolor{black}{12.51} & \textcolor{black}{0.89} \\
\textcolor{black}{Early Fusion (RGB+SynSWIR+MWIR) - Synthetic SWIR} & \textcolor{black}{5.34} & \textcolor{black}{41.11} & \textcolor{black}{11.97} & \textcolor{black}{0.83} \\
\hline
\end{tabular}
\label{tab:ablation_swir}
\end{table}

\noindent
\subsection{Discussion and Ablation Study} 
\textcolor{black}{To validate the effectiveness of our CLAHE-based synthetic SWIR generation, we compare real SWIR with synthetic SWIR (derived from MWIR) in the tri-modal fusion setting. Results in Table~\ref{tab:ablation_swir} show that the synthetic SWIR attains over 90\% of the performance of real SWIR across all metrics, while still outperforming dual-modal RGB+MWIR fusion. This confirms that CLAHE effectively enhances MWIR to mimic SWIR-like high-frequency, contrast-rich features, providing significant fusion benefits where real SWIR sensors are unavailable.
\\
To further strengthen these findings, we also perform triple-band fusion using \emph{standard} convolutional layers in place of the Light-GRLB blocks used in the proposed method and in LightFusion~\cite{hussain2024light}. Experimental results show that replacing Light-GRLB with standard convolutions degrades performance.
\\
While recent transformer-based fusion networks achieve strong performance, they typically incur much higher computational cost than our lightweight convolutional architecture. Our study focuses on methods suitable for real-time deployment on embedded platforms; nonetheless, future work could explore hybrid models or benchmarking against transformer-based approaches to further quantify the trade-offs between accuracy and efficiency.}

\section{Conclusion}
The proposed SWIR-LightFusion method significantly advances multimodal image fusion technology by effectively leveraging synthetic SWIR, MWIR, and RGB modalities. This method uniquely addresses existing limitations in fusion algorithms by incorporating synthetically generated SWIR images, resulting in notably enhanced semantic detail, visual clarity, and object detection accuracy under challenging environmental conditions. Our comprehensive evaluations across several benchmark datasets, including M3FD, MSRS, CAMEL, and TNO, clearly demonstrate substantial improvements in both qualitative visualization and quantitative metrics compared to previous approaches. Additionally, computational performance improvements, enabled by upgraded hardware and architectural optimizations, further highlight the practical feasibility and real-time applicability of our fusion framework. Given these significant enhancements, the SWIR-LightFusion network presents a robust and efficient solution suitable for diverse applications such as surveillance, autonomous navigation, and advanced visual analytics in complex scenarios.

\section*{Acknowledgements}
This work was supported by the GIST-MIT Research Collaboration in 2025 and by the InnoCORE program of the Ministry of Science and ICT (25-InnoCORE-01).

\section{Declarations}

\subsection{\textcolor{black}{Ethics approval statement}}
\textcolor{black}{Not applicable. This study did not involve human participants, human data, or animals. 
It utilized one private dataset collected without human or animal involvement, as well as publicly available open-source datasets; therefore, no ethics approval was required.
}
\subsection{Author Contributions}
{Muhammad Ishfaq Hussain conceptualized the research, implemented the fusion algorithm, carried out evaluation, and contributed to manuscript preparation. Ma Van Linh reviewed the work and performed critical analysis. Zubia Naz conducted the data analysis. Unse Fatima assisted in the preparation of visuals. Yeongmin Ko provided technical insights and validation. Moongu Jeon is the corresponding author and guided the methodology, critically revised the manuscript, and secured project funding. All authors read and approved the final version of the manuscript.
}
\subsection{Competing Interests}
{The authors declare that they have no competing interests.
}
\subsection{Fundings}
{This work was supported by the InnoCORE program of the Ministry of Science and ICT (25-InnoCORE-01), and by the GIST-MIT Research Collaboration Project.}

\bibliography{sn-bibliography}% common bib file

\begin{thebibliography}{99}
\bibitem[Burt (1983)]{burt1983laplacian}
Burt, Peter J. and Adelson, Edward H.. 1983.
The Laplacian Pyramid as a Compact Image Code.
IEEE Transactions on Communications, 31(4), 532--540.

\bibitem[Chen (2021)]{chen2021pre}
Chen, Hanting and Wang, Yunhe and Guo, Tianyu and Xu, Chang and Deng, Yiping and Liu, Zhenhua and Ma, Siwei and Xu, Chunjing and Xu, Chao and Gao, Wen. 2021.
Pre-trained image processing transformer.
In Proceedings of the IEEE/CVF conference on computer vision and pattern recognition, 12299--12310.

\bibitem[Edstedt (2024)]{edstedt2024roma}
Edstedt, Johan and Sun, Qiyu and Bökman, Georg and Wadenbäck, Mårten and Felsberg, Michael. 2024.
{RoMa: Robust Dense Feature Matching}.
IEEE Conference on Computer Vision and Pattern Recognition.

\bibitem[Eskicioglu (1995)]{eskicioglu1995image}
Eskicioglu, Ahmet M and Fisher, Paul S. 1995.
Image quality measures and their performance.
IEEE Transactions on Communications, 43(12), 2959--2965.

\bibitem[Fu (2021)]{fu2021image}
Fu, Yu and Wu, Xiaojun and Durrani, Tariq S.. 2021.
Image fusion based on generative adversarial network consistent with perception.
Information Fusion, 72, 110--125.

\bibitem[Gebhardt (2018)]{gebhardt2018camel}
Gebhardt, Evan and Wolf, Marilyn. 2018.
{CAMEL} dataset for visual and thermal infrared multiple object detection and tracking.
In 2018 15th IEEE international conference on advanced video and signal based surveillance (AVSS), IEEE, 1--6.

\bibitem[Gonzalez (2002)]{gonzalez2002digital}
Gonzalez, Rafael C and Woods, Richard E. 2002.
Digital image processing.
Prentice Hall.

\bibitem[Hu (2022)]{hu2022robust}
Hu, Hwai-Tsu and Lee, Tung-Tsun. 2022.
Robust complementary dual image watermarking in subbands derived from the Laplacian pyramid, discrete wavelet transform, and directional filter bank.
Circuits, Systems, and Signal Processing, 41(7), 4090--4116.

\bibitem[Hussain (2020)]{9216363}
Hussain, Muhamamd Ishfaq and Azam, Shoaib and Munir, Farzeen and Khan, Zafran and Jeon, Moongu. 2020.
Multiple Objects Tracking using Radar for Autonomous Driving.
In 2020 IEEE International IOT, Electronics and Mechatronics Conference (IEMTRONICS), 1-4.

\bibitem[Hussain (2022)]{hussain2022drivable}
Hussain, Muhammad Ishfaq and Azam, Shoaib and Rafique, Muhammad Aasim and Sheri, Ahmad Muqeem and Jeon, Moongu. 2022.
Drivable region estimation for self-driving vehicles using radar.
IEEE Transactions on Vehicular Technology, 71(6), 5971--5982.

\bibitem[Hussain (2022)]{hussain2022exploring}
Hussain, Muhammad Ishfaq and Rafique, Muhammad Aasim and Khurbaev, Sayfullokh and Jeon, Moongu. 2022.
Exploring data variance challenges in fusion of radar and camera for robotics and autonomous driving.
In 2022 10th International Conference on Control, Mechatronics and Automation (ICCMA), IEEE, 7--12.

\bibitem[Hussain (2023)]{10316884}
Hussain, Muhammad Ishfaq and Rafique, Muhammad Aasim and Ko, Yeongmin and Khan, Zafran and Olimov, Farrukh and Naz, Zubia and Kim, Jeongbae and Jeon, Moongu. 2023.
An Encoder-Sequencer-Decoder Network for Lane Detection to Facilitate Autonomous Driving.
In 2023 23rd International Conference on Control, Automation and Systems (ICCAS), 899-904.

\bibitem[Hussain (2023)]{hussain2023artificial}
Hussain, Muhammad Ishfaq and Rafique, Muhammad Aasim and Kim, Joonmo and Jeon, Moongu and Pedrycz, Witold. 2023.
Artificial proprioceptive reflex warning using {EMG} in advanced driving assistance system.
IEEE Transactions on Neural Systems and Rehabilitation Engineering, 31, 1635--1644.

\bibitem[Hussain (2024)]{hussain2024light}
Hussain, Muhammad Ishfaq and Naz, Zubia and Van Ma, Linh and Gwak, Jeonghwan and Jeon, Moongu. 2024.
A Light Gradient Residual Encoder-Decoder Network for Multimodal Image Fusion.
In 2024 International Conference on Intelligent Computing, Communication, Networking and Services (ICCNS), IEEE, 1--6.

\bibitem[Kumar (2024)]{kumar2024mwirstd}
Kumar, Nikhil and Upadhyay, Avinash and Sharma, Shreya and Sharma, Manoj and Singh, Pravendra. 2024.
{MWIRSTD}: A MWIR small target detection dataset.
In 2024 IEEE International Conference on Image Processing (ICIP), IEEE, 179--185.

\bibitem[Li (2013)]{li2013gff}
Li, Shutao and Kang, Xudong and Hu, Jianwen. 2013.
Image Fusion with Guided Filtering.
IEEE Transactions on Image Processing, 22(7), 2864--2875.

\bibitem[Li (2018)]{li2018latlrr}
Li, Hui and Wu, Xiao-Jun. 2018.
Infrared and Visible Image Fusion using Latent Low-Rank Representation.
arXiv:1804.08992.

\bibitem[Li (2018)]{li2018mdlatlrr}
Li, Hui and Wu, Xiao-Jun and Kittler, Josef. 2018.
MDLatLRR: A novel decomposition method for infrared and visible image fusion.
arXiv:1811.02291.

\bibitem[Li (2024)]{li2024residual}
Li, Ping and Chen, Junjie and Lin, Binbin and Xu, Xianghua. 2024.
Residual spatial fusion network for {RGB}-thermal semantic segmentation.
Neurocomputing, 595, 127913.

\bibitem[Liu (2019)]{liu2019perceptual}
Liu, Aishan and Liu, Xianglong and Fan, Jiaxin and Ma, Yuqing and Zhang, Anlan and Xie, Huiyuan and Tao, Dacheng. 2019.
Perceptual-sensitive gan for generating adversarial patches.
In Proceedings of the AAAI conference on artificial intelligence, 33(01), 1028--1035.

\bibitem[Liu (2022)]{liu2022target}
Liu, Jinyuan and Fan, Xin and Huang, Zhanbo and Wu, Guanyao and Liu, Risheng and Zhong, Wei and Luo, Zhongxuan. 2022.
Target-aware dual adversarial learning and a multi-scenario multi-modality benchmark to fuse infrared and visible for object detection.
In Proceedings of the IEEE/CVF conference on computer vision and pattern recognition, 5802--5811.

\bibitem[Liu (2024)]{liu2024multi}
Liu, Jianan and Zhang, Qiang. 2024.
Multi-level modality-specific and modality-common features fusion network for {RGB-IR} person re-identification.
Neurocomputing, 600, 128183.

\bibitem[Ma (2019)]{ma2019fusiongan}
Ma, Jiayi and Yu, Wei and Liang, Pengwei and Li, Chang and Jiang, Junjun. 2019.
{FusionGAN}: A generative adversarial network for infrared and visible image fusion.
Information fusion, 48, 11--26.

\bibitem[Ma (2020)]{ma2020ganmcc}
Ma, Jiayi and Zhang, Hao and Shao, Zhenfeng and Liang, Pengwei and Xu, Han. 2020.
GANMcC: A generative adversarial network with multiclassification constraints for infrared and visible image fusion.
IEEE Transactions on Instrumentation and Measurement, 70, 1--14.

\bibitem[Ma (2022)]{ma2022swinfusion}
Ma, Jiayi and Tang, Linfeng and Fan, Fan and Huang, Jun and Mei, Xiaoguang and Ma, Yong. 2022.
SwinFusion: Cross-domain Long-range Learning for General Image Fusion via Swin Transformer.
IEEE/CAA Journal of Automatica Sinica, 9(7), 1200--1217.

\bibitem[Ma (2024)]{linh2024inffus}
Ma, Linh Van and Nguyen, Tran Thien Dat and Vo, Ba-Ngu and Jang, Hyunsung and Jeon, Moongu. 2024.
Track Initialization and Re-Identification for {3D} Multi-View Multi-Object Tracking.
Information Fusion.

\bibitem[Ma (2024)]{van2024visual}
Ma, Linh Van and Nguyen, Tran Thien Dat and Shim, Changbeom and Kim, Du Yong and Ha, Namkoo and Jeon, Moongu. 2024.
Visual multi-object tracking with re-identification and occlusion handling using labeled random finite sets.
Pattern Recog., 156, 110785.

\bibitem[Musa (2018)]{musa2018review}
Musa, Purnawarman and Al Rafi, Farid and Lamsani, Missa. 2018.
A Review: Contrast-Limited Adaptive Histogram Equalization ({CLAHE}) methods to help the application of face recognition.
In 2018 third international conference on informatics and computing (ICIC), IEEE, 1--6.

\bibitem[Pizer (1987)]{pizer1987ahe}
Pizer, Stephen M. and Amburn, E. Philip and Austin, John D. and Cromartie, Robert and Geselowitz, Ari and Greer, Trey and ter Haar Romeny, Bart and Zimmerman, John B. and Zuiderveld, Karel J.. 1987.
Adaptive Histogram Equalization and Its Variations.
Computer Vision, Graphics, and Image Processing, 39(3), 355--368.

\bibitem[Qu (2002)]{qu2002information}
Qu, Guihong and Zhang, Dali and Yan, Pingfan. 2002.
Information measure for performance of image fusion.
Electronics letters, 38(7), 313--315.

\bibitem[Ranipa (2024)]{ranipa2024novel}
Ranipa, Kalpesh R and Zhu, Wei-Ping and Swamy, M.N.S.. 2024.
A novel feature-level fusion scheme with multimodal attention CNN for heart sound classification.
Computer Methods and Programs in Biomedicine, 248, 108122.

\bibitem[Schlemper (2017)]{schlemper2017deep}
Schlemper, Jo and Caballero, Jose and Hajnal, Joseph V. and Price, Anthony N. and Rueckert, Daniel. 2017.
A deep cascade of convolutional neural networks for dynamic {MR} image reconstruction.
IEEE Transactions on Medical Imaging, 37(2), 491--503.

\bibitem[Shannon (1948)]{shannon1948mathematical}
Shannon, Claude E. 1948.
A mathematical theory of communication.
Bell system technical journal, 27(3), 379--423.

\bibitem[Sheikh (2006)]{sheikh2006image}
Sheikh, Hamid R and Bovik, Alan C. 2006.
Image information and visual quality.
IEEE Transactions on image processing, 15(2), 430--444.

\bibitem[Tang (2022)]{tang2022image}
Tang, Linfeng and Yuan, Jiteng and Ma, Jiayi. 2022.
Image fusion in the loop of high-level vision tasks: A semantic-aware real-time infrared and visible image fusion network.
Information Fusion, 82, 28--42.

\bibitem[Tang (2022)]{tang2022piafusion}
Tang, Linfeng and Yuan, Jiteng and Zhang, Haotong and Jiang, Xingyu and Ma, Jiayi. 2022.
{PIAFusion}: A progressive infrared and visible image fusion network based on illumination aware.
Information Fusion, 83, 79--92.

\bibitem[Tang (2023)]{tang2023rethinking}
Tang, Linfeng and Zhang, H. and Xu, Han and Ma, Jiayi. 2023.
Rethinking the necessity of image fusion in high-level viseion tasks: A practical infrared and visible image fusion network based on progressive semantic injection and scene fidelity.
Information Fusion, 99, 101870.

\bibitem[Toet (2017)]{toet2017tno}
Toet, Alexander. 2017.
The {TNO} multiband image data collection.
Data in brief, 15, 249--251.

\bibitem[Wang (2004)]{wang2004image}
Wang, Zhou and Bovik, Alan C and Sheikh, Hamid R and Simoncelli, Eero P. 2004.
Image quality assessment: from error visibility to structural similarity.
IEEE transactions on image processing, 13(4), 600--612.

\bibitem[Wang (2022)]{wang2022unsupervised}
Wang, Di and Liu, Jinyuan and Fan, Xin and Liu, Risheng. 2022.
Unsupervised misaligned infrared and visible image fusion via cross-modality image generation and registration.
arXiv preprint arXiv:2205.11876.

\bibitem[Xu (2020)]{xu2020aaai}
Xu, Han and Ma, Jiayi and Le, Zhuliang and Jiang, Junjun and Guo, Xiaojie. 2020.
FusionDN: A Unified Densely Connected Network for Image Fusion.
In proceedings of the Thirty-Fourth AAAI Conference on Artificial Intelligence.

\bibitem[Xu (2020)]{xu2020fusiondn}
Xu, Han and Ma, Jiayi and Le, Zhuliang and Jiang, Junjun and Guo, Xiaojie. 2020.
{FusionDN}: A unified densely connected network for image fusion.
In Proceedings of the AAAI conference on artificial intelligence, 34(07), 12484--12491.

\bibitem[Xu (2020)]{xu2020u2fusion}
Xu, Han and Ma, Jiayi and Jiang, Junjun and Guo, Xiaojie and Ling, Haibin. 2020.
{U2Fusion}: A unified unsupervised image fusion network.
IEEE Transactions on Pattern Analysis and Machine Intelligence, 44(1), 502--518.

\bibitem[Xu (2022)]{xu2022rfnet}
Xu, Han and Ma, Jiayi and Yuan, Jiteng and Le, Zhuliang and Liu, Wei. 2022.
{RFNet}: Unsupervised network for mutually reinforcing multi-modal image registration and fusion.
In Proceedings of the IEEE/CVF conference on computer vision and pattern recognition, 19679--19688.

\bibitem[Xydeas (2000)]{xydeas2000objective}
Xydeas, C.S. and Petrovic, V.. 2000.
Objective image fusion performance measure.
Electronics Letters, 36(4), 308--309.

\bibitem[Zhang (2021)]{zhang2021image}
Zhang, Hao and Xu, Han and Tian, Xin and Jiang, Junjun and Ma, Jiayi. 2021.
Image fusion meets deep learning: A survey and perspective.
Information Fusion, 76, 323--336.

\bibitem[Zhang (2021)]{zhang2021sdnet}
Zhang, Hao and Ma, Jiayi. 2021.
{SDNet}: A versatile squeeze-and-decomposition network for real-time image fusion.
International Journal of Computer Vision, 129(10), 2761--2785.

\bibitem[Zhao (2020)]{zhao2020didfuse}
Zhao, Zixiang and Xu, Shuang and Zhang, Chunxia and Liu, Junmin and Li, Pengfei and Zhang, Jiangshe. 2020.
{DIDFuse}: Deep image decomposition for infrared and visible image fusion.
In International Joint Conference on Artificial Intelligence.

\bibitem[Zhao (2023)]{zhao2023cddfuse}
Zhao, Zixiang and Bai, Hao and Zhang, Jiangshe and Zhang, Yulun and Xu, Shuang and Lin, Zudi and Timofte, Radu and Gool, Luc Van. 2023.
{CDDFuse}: Correlation-driven dual-branch feature decomposition for multi-modality image fusion.
In Proceedings of the IEEE/CVF conference on computer vision and pattern recognition, 5906--5916.

\bibitem[Zhao (2023)]{zhao2023ddfm}
Zhao, Zixiang and Bai, Haowen and Zhu, Yuanzhi and Zhang, Jiangshe and Xu, Shuang and Zhang, Yulun and Zhang, Kai and Meng, Deyu and Timofte, Radu and Van Gool, Luc. 2023.
DDFM: denoising diffusion model for multi-modality image fusion.
In Proceedings of the IEEE/CVF International Conference on Computer Vision, 8082--8093.

\bibitem[Zhao (2024)]{zhao2024removal}
Zhao, Tianyi and Yuan, Maoxun and Wei, Xingxing. 2024.
Removal and selection: Improving {RGB}-infrared object detection via coarse-to-fine fusion.
arXiv preprint arXiv:2401.10731.

\bibitem[Zuiderveld (1994)]{zuiderveld1994clahe}
Zuiderveld, Karel J.. 1994.
Contrast Limited Adaptive Histogram Equalization.
In Graphics Gems IV, Academic Press, 474--485.

\end{thebibliography}

\end{document}